\pdfoutput=1

\documentclass[11pt]{article}

\usepackage{acl}
\usepackage{listings}
\usepackage{times}
\usepackage{latexsym}
\usepackage{booktabs}
\usepackage[T1]{fontenc}
\usepackage{fdsymbol}
\usepackage[utf8]{inputenc}

\usepackage{inconsolata}

\usepackage{graphicx}
\usepackage{cleveref}

\title{When Does Meaning Backfire? Investigating the Role of AMRs in NLI} 

\author{
    Junghyun Min\textsuperscript{$\smwhitestar$} \quad
    Xiulin Yang\textsuperscript{$\smwhitestar$}  \quad
    Shira Wein\textsuperscript{$\smwhtdiamond$}
    \\[1ex]
    \textsuperscript{$\smwhitestar$}Georgetown University \quad
    \textsuperscript{$\smwhtdiamond$}Amherst College
    \\
    {
        \texttt{\{jm3743, xy236\}@georgetown.edu} \quad
        \texttt{swein@amherst.edu }
    }
}

\begin{document}
\maketitle
\begin{abstract}
Natural Language Inference (NLI) relies heavily on adequately parsing the semantic content of the premise and hypothesis.
In this work, we investigate whether adding semantic information in the form of an Abstract Meaning Representation (AMR) helps pretrained language models better generalize in NLI. 
Our experiments\footnote{We publicly release our code at \url{https://github.com/Aatlantise/advarsarial-nli-amr}.} integrating AMR into NLI in both fine-tuning and prompting settings show that the presence of AMR in fine-tuning hinders model generalization while prompting with AMR leads to slight gains in \texttt{GPT-4o}.
However, an ablation study reveals that the improvement comes from amplifying surface-level differences rather than aiding semantic reasoning. 
This amplification can mislead models to predict non-entailment even when the core meaning is preserved.

\end{abstract}

\section{Introduction}
\label{sec:intro}
Since the advent of large language models (LLMs), there has been ongoing debate about the utility of symbolic representations such as Abstract Meaning Representations (AMRs; \citealp{banarescu-etal-2013-abstract}) in (LLM-based) pipelines and existing NLP tasks. While some studies report limited or negative impact of AMRs on mainstream NLP tasks \citep{jin-etal-2024-analyzing}, recent work has demonstrated their value in specific applications, such as syntactic simplification \citep{yao-etal-2024-semantic} and semantically-controllable text transformation \citep{li2025sentence}. Perhaps unsurprisingly, incorporating AMR has been particularly well-explored and effective in tasks related to semantics \citep{wein-opitz-2024-survey}.

\begin{figure}
    \centering
    \includegraphics[width=\linewidth]{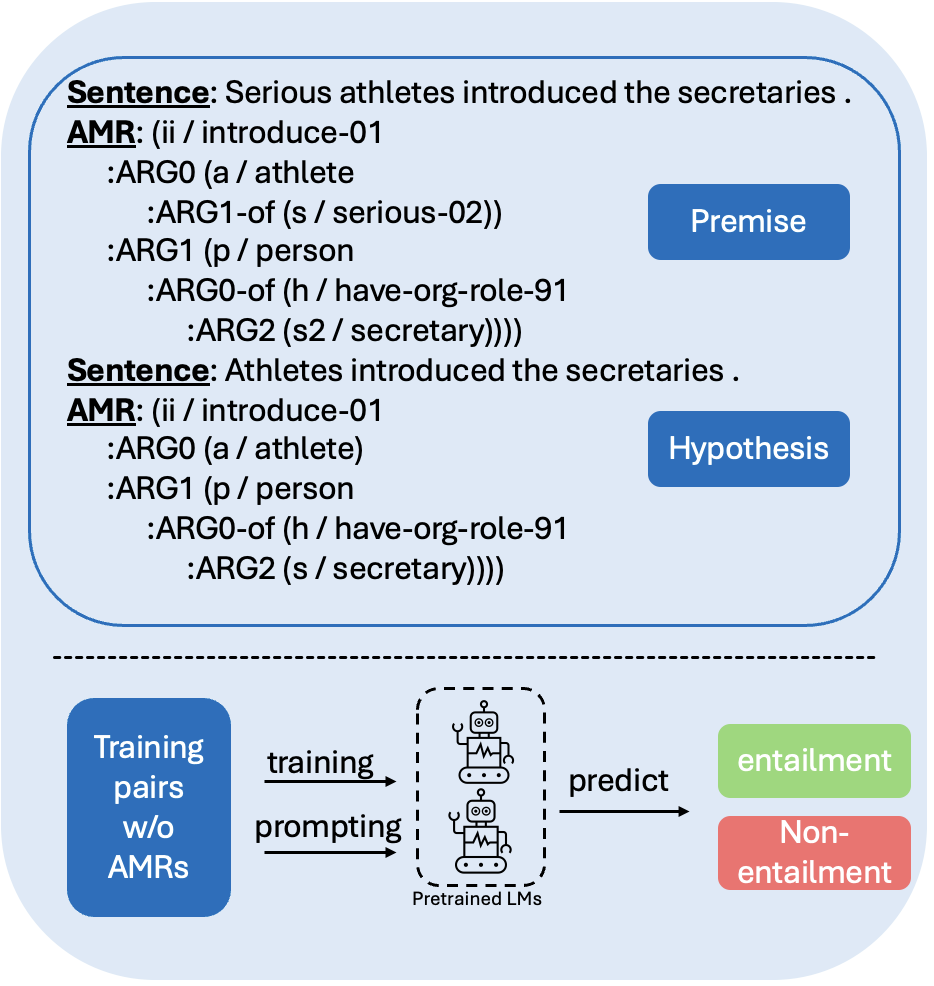}
    \caption{An example of NLI and experiment pipeline of the paper. AMRs are shown in penman notation. 
    }
    \label{fig:exp}
\end{figure}

Natural language inference \citep[NLI; ][]{DAGAN_2010_RECOGNIZING} is a popular task in NLP where the solver is given a \textit{premise} and a \textit{hypothesis}, and asked to determine whether the hypothesis is true if the premise is true. The label space consists of three labels: \textbf{entailment} if the hypothesis is true, \textbf{contradiction} if the hypothesis is false, and \textbf{neutral} if the truth value of the hypothesis cannot be determined; this can also be condensed in two labels: entailment and non-entailment.
As shown in \Cref{fig:exp} ``Athletes introduced the secretaries'' should be entailed by ``Serious athletes introduced the secretaries.'' Therefore, the label should be \textbf{entailment} because the truth of the premise indicates truth of (or \emph{entails}) the hypothesis.

As a meaning-focused task, NLI aligns well with the motivation behind AMRs, i.e., to abstract sentence meaning beyond surface form, given NLI models' tendencies to adopt shallow heuristics rather than understanding the relationship between the premise and the hypothesis, leading to poor generalization to novel data \citep{gururangan-etal-2018-annotation, poliak-etal-2018-hypothesis, mccoy-etal-2019-right, serrano-etal-2023-stubborn}. In this paper, we investigate whether incorporating AMRs as additional input - either during (a) fine-tuning or (b) prompting - can encourage models to attend more to abstract meaning, thereby improving generalization and overall performance.  As illustrated in \Cref{fig:exp}, we add AMRs to either the training data or prompts then evaluate how the addition of AMR affects generalization performance. We find that AMRs generally hinder performance in both fine-tuning and prompting settings, with the exception of prompting on HANS. However, this improvement appears to stem from AMRs amplifying surface-level differences rather than capturing deeper semantic meaning.




\section{Related Work}

NLI \citep{DAGAN_2010_RECOGNIZING}
is a hallmark task demonstrating model's ability to understand natural language.
Select neural models like \texttt{BERT} \citep{devlin-etal-2019-bert} and \texttt{RoBERTa} \citep{liu2019roberta} trained on datasets like Multi-genre NLI \citep[MNLI; ][]{williams-etal-2018-broad} and Stanford NLI \citep[SNLI; ][]{bowman-etal-2015-large} provide test-set performance close to that of humans \citep{nangia-bowman-2019-human}, but the near-human performance on MNLI has been attributed to models optimizing on the spurious correlations between lexical items and labels in the data \citep{poliak-etal-2018-hypothesis, mccoy-etal-2019-right, gururangan-etal-2018-annotation, serrano-etal-2023-stubborn}. The same models that excel in test-set performance suffer from poor generalization to other datasets that represent the same task \citep{zhou-etal-2020-curse, mccoy-etal-2020-berts, delbari2025beyond}.

Several prior approaches have incorporated logical representations into NLI, for example by combining neural encoders with logical reasoning modules \citep{chen-etal-2021-neurallog}, training natural logic theorem provers \citep{abzianidze-2020-learning}, extracting phrase correspondences via natural deduction proofs \citep{yanaka-etal-2018-acquisition}, or constraining large language models with natural logic inference patterns \citep{noble-etal-2025-mood}. While these works rely on task-specific inference rules or specialized proof systems, our use of AMRs differs in that AMRs provide a broad, task-agnostic semantic abstraction without requiring dedicated engineering. Once an AMR parser is available, AMRs can be used as direct inputs to pretrained models such as BERT (Section \ref{sec:exp-fine-tune} and ChatGPT (Section \ref{sec:exp-llms}), enabling structured input with minimal task-specific engineering.

LLMs and in-context learning have been used to tackle NLI and generalization in it, with mixed results; \citet{webson-pavlick-2022-prompt} show that the content of prompts do not significantly influence LLMs' performance in NLI tasks, while \citet{kavumba-etal-2023-prompting, he-etal-2024-using} use chain-of-thought reasoning and natural language explanations to improve NLI performance and generalization. However, \citet{zhong2023can} report that its NLI performance is still only comparable to much smaller encoder-only models like \texttt{BERT} and \texttt{RoBERTa} \citep{devlin-etal-2019-bert, liu2019roberta}, leaving adversarial NLI an ongoing area of research.


Recent work on AMRs has set out to utilize AMR graphs for a variety of downstream tasks, including summarization and information extraction (see \citet{wein-opitz-2024-survey,sadeddine-etal-2024-survey} for comprehensive overviews). AMRs excel in capturing structure-dependent meaning \citep{leung-etal-2022-semantic} and have shown particular promise in meaning-sensitive tasks such as debiasing translationese \citep{wein-schneider-2024-lost}, style transfer \citep{hua-etal-2023-improving},  and sentence-level manipulation \citep{li2025sentence}, especially when used in conjunction with fine-tuned models. 

To the best of our knowledge, \citet{opitz-etal-2023-amr4nli} represents the only prior effort to incorporate AMRs into NLI, and do so for the purpose of interpretable NLI evaluation. They find that metrics based on AMR are robust unsupervised representations of premise-hypothesis relationships when used alongside neural representations like \texttt{BERT}. 


\section{Data \& Experiments}


\subsection{Data \& Models}
In these experiments, we use two datasets: MNLI \citep{williams-etal-2018-broad} and HANS \citep{mccoy-etal-2019-right}. 
MNLI is a crowdsourced dataset, with a test set that is not available to the public. We follow prior work \citep{wang-etal-2018-glue, devlin-etal-2019-bert} in taking one of its two developmental splits as the evaluation dataset. Specifically, we take the \texttt{matched} developmental set to use as our evaluation dataset. The training dataset includes 297k sentence pairs, while the evaluation set contains around 10k pairs.
HANS is a template-based evaluation dataset, with 30k examples. 
Unlike MNLI and other NLI datasets, its label space consists of only two labels--\texttt{entailment} and \texttt{non-entailment}. 
We follow prior work \citep{mccoy-etal-2020-berts, min-etal-2020-syntactic} in collapsing the model's \texttt{neutral} and \texttt{contradiction} predictions to the single \texttt{non-entailment} label when calculating evaluation metrics, to accommodate the two-class label space of HANS.


We use an off-the-shelf AMR parser from \texttt{amrlib} \footnote{\url{https://github.com/bjascob/amrlib-models/releases/tag/parse_xfm_bart_large-v0_1_0}} to parse all the sentences from the two datasets we use. The model is \texttt{BART-large} \citep{lewis2019bart} fine-tuned on AMR 3.0 \citep{knight2021abstract}. 
While parsers with higher reported scores exist \citep[e.g. ][]{bevilacqua-etal-2021-one}, we follow \citet{uhrig-etal-2021-translate, opitz-etal-2023-amr4nli} in selecting an \texttt{amrlib} parser for ease of implementation.

We manually perform a small sanity check over a subset of generated AMRs to verify that AMR parses are acceptable, but do not perform a comprehensive quality check over the entire dataset. We observe that the AMRs produced for sentences in the HANS dataset are generally acceptable, likely benefiting from the sentences' simple structure and short length, though certainly the generated AMRs contain noise; the sentences in MNLI are longer and more complex.



\subsection{Experiment 1: Can fine-tuned models benefit from AMR in NLI?}
\label{sec:exp-fine-tune}

We train three sets of \texttt{BERT-base} models, augmented with AMR information to perform our experiment. We incorporate AMR in three ways:
(1)~linearized AMR is concatenated to text input (+AMR as text); (2)~graph neural network representation of AMR is concatenated to text representation (+AMR as graph); and (3)~just the linearized AMR is used as text input (AMR as text only).

We adopt the setup and hyperparameters of previous work in MNLI fine-tuning and HANS evaluation \citep{mccoy-etal-2020-berts, min-etal-2020-syntactic}. 
We take the \texttt{bert-base-uncased} model and fine-tune for 3 epochs with a learning rate of 2e-5. 
While we opt to follow prior work, we note that longer fine-tuning beyond 10 epochs at the same learning rate significantly improves HANS performance in all settings. 
Each label prediction is made from the \texttt{[CLS]} token's final layer embedding. While the setup is equivalent to those from prior work, we implement the setup in a more modern, current stack. 
Due to updates in the hardware and software since prior work, slight changes in the resulting model weights are possible.
To control for such an effect, we perform a sanity check via baseline in-distribution test set evaluation.
Finally, we integrate AMR into the models as text, via \emph{linearization}, removing all newlines and whitespace sequences longer than length two.

\subsection{Experiment 2: Can prompt-based models benefit from AMR in NLI?}
\label{sec:exp-llms}
In this experiment, we evaluate whether incorporating AMRs improves LLMs' performance on NLI, on both the MNLI and HANS dataset, the latter of which remains challenging even after fine-tuning. \citet{jin-etal-2024-analyzing} find that only instruction-tuned GPT models are capable of reliably processing AMRs. We therefore restrict our evaluation to \texttt{GPT-4o} \citep{hurst2024gpt} in zero-shot and 5-shot settings. 

We use the following prompt template:
\begin{quote}
    \textit{You are a helpful assistant trained to determine whether a hypothesis logically follows from a premise. Respond with 'Yes' or 'No'.\\
    Premise: [X].\\
    Hypothesis: [Y].}
\end{quote}

Where [X] and [Y] are replaced with the premise and hypothesis in question.  The prompt applies to both zero and few-shot settings.
We incorporate no additional details or explanations about the task (NLI), the datasets (MNLI and HANS), or the AMRs in our prompt, to best measure the LLM's ability to use representations of meaning for NLI, rather than perform in-context learning.
However, it is possible that the model perform better with additional context on the task, dataset, or AMRs.

We test three input conditions: (a) sentence only; (b) AMR only; and (c) sentence + AMR. Label preprocessing follows the same procedure as in the fine-tuning setup for MNLI. In the 5-shot setting, we randomly sampled 5 examples from the training set of each data set. We set the temperature to 0 to ensure deterministic outputs.

\begin{table}[t]
    \centering
    \small
    \begin{tabular}{l|cc}
    \toprule
        \textbf{Model} & \textbf{MNLI} & \textbf{HANS}  \\
        
        \midrule
        Chance & 0.33 & 0.50 \\
        Baseline \citep{mccoy-etal-2020-berts} & 0.84 & 0.57 \\ 
        \quad +Syntactic aug \citep{min-etal-2020-syntactic} & 0.84 & 0.65 \\
        \midrule
        \textbf{Ours} \\
        Baseline reproduction (text only) & 0.84 & 0.52 \\
        \quad +AMR as text & 0.83 & 0.47 \\
        \quad +AMR as graph & 0.84 & 0.49 \\
        AMR as text only & 0.74 & 0.51 \\
        \bottomrule
    \end{tabular}
    \caption{Performance comparison with and without AMR on HANS and MNLI test sets in the fine-tuning setting. Both datasets measure accuracy.}
    \label{tab:fine-tune-results}
\end{table}

\section{Results \& Discussion}
\subsection{Experiment 1}
We report the accuracies of our fine-tuning models with and without AMRs in \Cref{tab:fine-tune-results}. We report numbers from prior work \citep{mccoy-etal-2020-berts, min-etal-2020-syntactic} in addition to our experiments to serve as comparison baselines and to ensure our setup is correct. Our reported numbers are an average across 10 runs with varying seed.

As shown in \Cref{tab:fine-tune-results}, AMR augmentation does not yield improvements in MNLI performance, nor HANS generalization. Perhaps analogously to previous data-driven attempts at improving generalization \citep{clark-etal-2019-dont, min-etal-2020-syntactic, yaghoobzadeh-etal-2021-increasing}, additional AMR information as either text or graph does not affect MNLI performance. Analysis of their confusion matrices reveals AMR adds or subtracts little in terms of MNLI label decision boundary. On HANS performance, We discuss two main findings.

\paragraph{Standalone AMR input for classification intensifies heuristics favoring the entailment label.}

AMR-only models predict the entailment label for 98.3\% of HANS examples, compared to the baseline models at 94.7\%. We attribute this to an intensified version of the baseline models' heuristic correlating overlap between the hypothesis and premise to the entailment label, dubbed the lexical overlap heuristic \citep{mccoy-etal-2019-right}. We note this is concurrent with a still competitive MNLI performance, at 84\%. We discuss this phenomenon in more detail in \Cref{sec:error-analysis-overlap,sec:error-analysis-combination}.

\paragraph{Mixing AMRs and text leads to more (false) negative predictions in novel data.}

On the other hand, combining AMR information with text strongly affects HANS label decision boundaries in the opposite direction, overriding various shallow heuristics that favor the entailment label observed in \citet{mccoy-etal-2020-berts} and in our baseline and AMR-only experiments. Our \texttt{+AMR as text} models 86.6\% of HANS examples, and \texttt{+AMR as graph} models 86.9\%, even predicting non-entailment on highly overlapping examples. We attempt to disentangle the effects of AMRs and text in a combined representation in \Cref{sec:error-analysis-combination}, where we find that while AMR can be used to perform NLI, it is less effective than text input and combining the two introduces new artifacts that are more difficult to interpret.

\subsection{Experiment 2}
\begin{table}
\small
\centering
\begin{tabular}{l|cc}
\toprule
Model & MNLI & HANS \\
\midrule
Chance & 0.33 & 0.50 \\
ChatGPT-3.5 \\
\quad \citet{zhong2023can} & 0.89 & - \\
\quad \citet{he-etal-2024-using} & - &  0.75 \\
\midrule
\textbf{Ours (ChatGPT-4o)} \\
Text only & \textbf{0.91} & 0.82 \\
\quad +AMR & 0.75 & \textbf{0.87} \\
AMR only & 0.68 & 0.70 \\


\bottomrule
\end{tabular}
\caption{Performance comparison with and without AMR on HANS and MNLI test sets in the LLM zero-shot prompting setting. 
}
\label{tab:amr-results}
\end{table}

The results for prompting with \texttt{GPT-4o} are shown in \Cref{tab:amr-results}. We report only the zero-shot results in the main text, as they yield similar overall performance and prediction patterns. Results for the five-shot setting are provided in \Cref{5shot}. Two main observations emerge. 

\paragraph{AMRs increase (false) negative predictions.} As shown in the table, model performance is consistently lowest when prompted with AMRs alone, while including the original sentence improves results.  We find this is because AMRs lead models to make more negative predictions (see \Cref{llm-stats}). To test this statistically, we fit a logistic regression model predicting non-entailment using SMATCH++ \citep{opitz-2023-smatch} between hypothesis and premise AMRs and data source (gold vs. predicted). A significant negative interaction ($\beta$ = -0.042, $p$ < 2e-16) shows that SMATCH similarity influences model predictions more than gold labels.

\begin{figure}
\centering
\includegraphics[width=\linewidth]{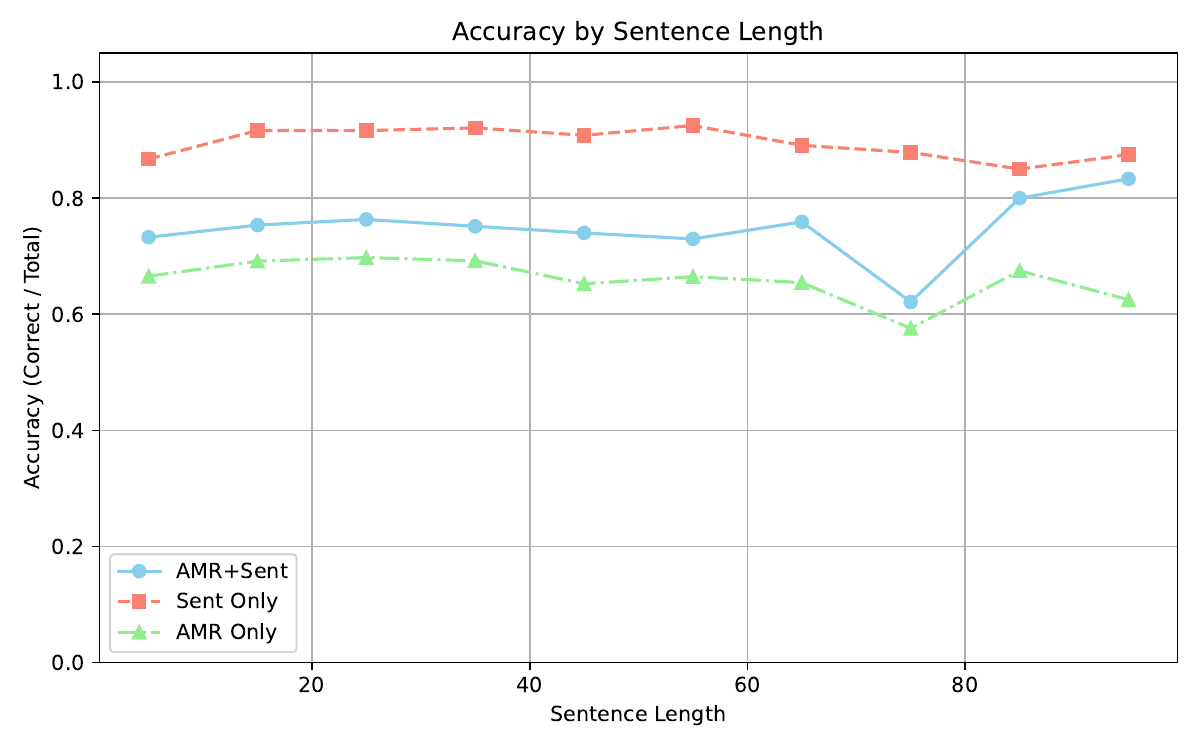}
\caption{Accuracy of three prompt settings across different sentence lengths on MNLI. 
}
\label{fig:acc}
\end{figure}
Further analysis reveals that AMR’s sensitivity to surface-level lexical and syntactic variation leads to low structural overlap between semantically equivalent expressions,\footnote{See \Cref{error_example} for an example.} misleading the model toward non-entailment. This also explains why, on the HANS test set, prompts that include both the sentence and its AMR lead to the highest rate of negative predictions: the AMR representation amplifies subtle differences between two otherwise similar strings, making semantic mismatches more salient and pushing the model toward rejecting entailment. Such nuanced contrasts are what HANS is designed to probe in language models, prompting \texttt{GPT-4o} to overpredict non-entailment.

\paragraph{AMR does not lead to more robust performance with longer sequence length.}

\citet{opitz-etal-2023-amr4nli} reported that incorporating AMRs improves robustness in NLI prediction. We investigate whether this finding holds for LLMs. Specifically, we plot accuracy across NLI examples binned by total sequence length (premise + hypothesis). For sequences exceeding 100 words, we group them into a single bin due to their sparsity. 

As shown in \Cref{fig:acc}, when \texttt{GPT-4o} is prompted with both sentence and AMR inputs, accuracy slightly increases for inputs longer than 80 words. However, this performance remains lower than that of sentence-only prompts across most length bins. We find no evidence that AMR-only prompts enhance robustness to longer sequences.

\subsection{Summary}
Our fine-tuning experiments suggest that AMR-only models are still susceptible to heuristics. We also observe that combining text with AMR as both graph and text is challenging and results in a strong preference towards the non-entailment label, even for highly overlapping, entailing examples.

Our LLM experiments showcase similar preference towards the non-entailment label. This suggests that AMRs effectively highlight subtle distinctions between minimal pairs, explaining improved \textbf{HANS} performance. However, for simpler examples, this heightened contrast can cause the model to overpredict \textit{No}, even for entailing sentence pairs.

\section{Conclusion}

In this work, we investigate whether AMRs can help PLMs on the task of natural language inference. Specifically, across both fine-tuning and prompting settings, we evaluate whether incorporating AMRs improves entailment classification.

We find that our implementations of AMR integration does not improve performance in fine-tuning, and only lead to slight gains in zero-shot prompting with \texttt{GPT-4o}. Importantly, ablation analyses reveal that these gains are not due to deeper semantic understanding, but rather to AMRs exaggerating surface-level differences, which in some cases mislead the model to predict \textit{non-entailment} where entailment holds. Overall, our results suggest that while AMRs offer a promising abstraction mechanism, their integration with LLMs requires careful design to avoid reinforcing shallow heuristics rather than promoting robust reasoning.

\section*{Limitations and Future Work} 
This study focuses on two datasets (MNLI and HANS) and explores a limited set of prompting and fine-tuning configurations. For fine-tuning, we adopt a single AMR linearization strategy; in the prompting setting, we test one prompt template with different conditions. While alternative prompts for zero-shot inference may yield better performance \citep[e.g.,][]{kavumba-etal-2023-prompting}, our consistent experimental setup enables fair comparisons across conditions. Nonetheless, the findings may not generalize to other inference tasks, domains, or prompting strategies.

Future work could explore more diverse linearization formats, prompt designs, and integration strategies that align AMR structure more directly with model attention or reasoning processes. 

Encoder-based models have been shown to be sensitive to minor perturbations in input \citep{sinha-etal-2021-perturbing, Jin_Jin_Zhou_Szolovits_2020}, and prior work integrating AMR graphs into neural models have used a variety of formats \citep{wein-opitz-2024-survey}. 
Thus, in addition to Python-like and natural language-like representation of AMR's structure \citep{srivastava2025instruction, srivastava2025revisiting, dutt-etal-2025-dependency-parses}, carefully designing how hierarchical devices in AMRs (e.g. variable names, parentheses, indents, and newlines) could be represented in the embedding space of encoder-only models may be worth further investigations.

Finally, investigating how AMRs interact with LLM decoding beyond surface augmentation may help unlock their full potential in meaning-sensitive tasks.

\section*{Responsible Research Statement}
We use \texttt{ChatGPT-4o} \citep{hurst2024gpt} as a coding assistant during the implementation of our experiments, in addition to as a natural language processor.

\section*{Acknowledgments}
This work was developed on a course project for Advanced Semantic Representation at Georgetown University. We thank Nathan Schneider and our classmates from the course for their helpful comments. A portion of experiments was conducted on the \texttt{gcp-gu-hpc-cli} cluster within the High Performance Computing at Georgetown University infrastructure.

\bibliography{custom}

\appendix

\section{Fine-tuning Error Analyses}
\subsection{Intensified subsequence overlap heuristic with AMR}
\label{sec:error-analysis-overlap}



Compared to text-only MNLI models which are known to incorrectly correlate lexical and sequence overlap to the entailment label \citep{mccoy-etal-2019-right, mccoy-etal-2020-berts, min-etal-2020-syntactic}, the AMR-only models favor the entailment label even more. The model's preference toward the entailment label results in the AMR-only models consistently predicting non-entailment for around 98\% of HANS examples. 

It is less likely that the model adopts the subsequence and constituency heuristic, as no text subsequences or constituencies are provided in the training dataset--only AMR parses are provided as input. However, it is possible that a new heuristic had formed. Consider the following two versions of premise-hypothesis pairs:

\begin{itemize}
    \item \noindent\textbf{Premise:} \textit{The judge and the president advised the scientist.}
\begin{lstlisting}[language=]
(a / advise-01
      :ARG0 (a2 / and
            :op1 (p / person
                  :ARG0-of (h / have-org-role-91
                        :ARG3 (j / judge-01)))
            :op2 (p2 / person
                  :ARG0-of (h2 / have-org-role-91
                        :ARG2 (p3 / president))))
      :ARG1 (s / scientist))
\end{lstlisting}

    \item \noindent\textbf{Hypothesis 1, label=non-entailment:} \textit{The scientist advised the judge.}
\begin{lstlisting}[language=]
(a / advise-01
      :ARG0 (s / scientist))
      :ARG1 (p / person
            :ARG0-of (h / have-org-role-91
                  :ARG3 (j / judge-01)))
\end{lstlisting}

    \item \noindent\textbf{Hypothesis 2, label=entailment:} \textit{The judge advised the scientist.}
\begin{lstlisting}[language=]
(a / advise-01
      :ARG0 (p / person
            :ARG0-of (h / have-org-role-91
                  :ARG3 (j / judge-01)))
      :ARG1 (s / scientist))
\end{lstlisting}
\end{itemize}

The premise and hypothesis AMRs exhibit significant overlap, namely in variables \texttt{p, h, j, s}. Given sufficiently many pairs similar to \texttt{Premise-Hypothesis 2} in the training set, the model may optimize to correlate variable overlap to the entailment label. Then, when the model predicts on \texttt{Premise-Hypothesis 1} pair, instead of considering the semantic structure, it may attend to the significant variable overlap, and predict \texttt{entailment}, which is the incorrect answer.

\subsection{Cross-setting evaluation analysis}
\label{sec:error-analysis-combination}

To disentangle the effects of text and AMR in \texttt{+AMR} models, we evaluate models in not only their own evaluation setting, but in other settings as well. We do not evaluate all models on all settings. Instead, we measure performance on reasonable train-evaluation setting pairs--we do not evaluate \texttt{text only} models on \texttt{AMR only} settings, and vice versa. \texttt{+AMR} models undergo all evaluation settings; all models undergo evaluation in the \texttt{+AMR} setting. In this cross-setting evaluation scheme, we do not consider the \texttt{+AMR as graph} setting.

\begin{table}
    \centering
    \small
    \begin{tabular}{c|ccc}
        \toprule
        Train setting & Text eval & +AMR eval & AMR only eval \\
        \midrule
        Text & 0.84 & 0.47 \\
        +AMR & 0.53 & 0.83 & 0.36 \\
        AMR only & & 0.44 & 0.74\\
        \bottomrule
    \end{tabular}
    \caption{MNLI accuracy of our trained models evaluated on each setting. Chance performance is 0.33.}
    \label{tab:mnli-cross-setting}
\end{table}

\begin{table}
    \centering
    \small
    \begin{tabular}{c|ccc}
        \toprule
        Train setting & Text eval & +AMR eval & AMR only eval \\
        \midrule
        Text & 0.96 & 0.97 \\
        +AMR & 0.25 & 0.13 & 0.85 \\
        AMR only & & 0.99 & 0.98\\
        \bottomrule
    \end{tabular}
    \caption{Percentage of HANS examples where our trained models evaluated on each setting predict entailment.}
    \label{tab:hans-ent-cross-setting}
\end{table}

First, we observe that cross-evaluation models still perform above chance in MNLI evaluation (0.33), as seen in \Cref{tab:mnli-cross-setting}, which indicates both text and AMR knowledge can be leveraged despite noise from unseen form. The \texttt{+AMR} models' single-mode (text only or AMR only) MNLI accuracies, together with the lower performance of \texttt{AMR-only} models compared to \texttt{text-only} models indicate that AMR information is more difficult to acquire and use than text input.

Second, we observe that bias towards entailment or non-entailment in MNLI and HANS is strongly correlated, given train-evaluation mismatch ($m = 0.6$, $R^2 = 0.87$). Cross-setting evaluation results support the case of a newly developed heuristic for \texttt{AMR only} models, as single-mode models overwhelmingly predict entailment in HANS examples even when evaluated on \texttt{+AMR} settings, both at above 96\%, as seen in \Cref{tab:hans-ent-cross-setting}.

On the other hand, it is difficult to pinpoint the cause of the tendency to predict non-entailment in dual-mode models predicting on input containing text. We observe that dual-mode models predict non-entailment for entailing adverbial sentences whose AMRs highly overlap, as shown below:

\begin{itemize}
    \item \noindent\textbf{Premise:} \textit{Clearly the bankers waited.}
\begin{lstlisting}[language=]
(w / wait-01 
    :ARG1 (b / banker) 
    :ARG1-of (c / clear-06))
\end{lstlisting}

    \item \noindent\textbf{Hypothesis, label=entailment, pred=non-entailment:} \textit{The bankers waited.}
\begin{lstlisting}[language=]
(w / wait-01 
    :ARG1 (b / banker) 
\end{lstlisting}
\end{itemize}

\section{LLM prediction statistics}
The results are reported in \Cref{tab:negative-count}. 
\label{llm-stats}
\begin{table}
\small
\centering
\begin{tabular}{l|cc}
\toprule
Prompt & MNLI & HANS \\
\midrule
Text + AMR & +2,274 & +1,759 \\
AMR only & +2,848 & +1,393 \\
\bottomrule
\end{tabular}
\caption{Increase in the number of negative predictions compared to the sentence-only prompt condition.}
\label{tab:negative-count}
\end{table}

\section{LLM setting error analysis: Example}
\label{error_example}

For example, while the premise \textit{everything you're looking for is available} is semantically equivalent to the hypothesis \textit{everything can be found}, the AMRs for these sentences differ substantially due to lexical choices (e.g., \textit{look for} vs. \textit{find}) and syntactic voice (active vs. passive). The resulting SMATCH++ F-score \citep{opitz-2023-smatch} between the two graphs is only 27.7.

\begin{itemize}
    \item \noindent\textbf{Premise:} \textit{Enter the realm of shopping malls, where everything you're looking for is available without moving your car.}
\begin{lstlisting}[language=]
(e / enter-01
      :ARG0 (y / you)
      :ARG1 (r / realm
            :mod (m / mall
                  :mod (s / shop-01)
                  :location-of (a / available-02
                        :ARG2 (e2 / everything
                              :ARG1-of (l / look-01
                                    :ARG0 y))
                        :manner (m2 / move-01
                              :polarity -
                              :ARG0 y
                              :ARG1 (c / car
                                    :poss y))))))
\end{lstlisting}

    \item \noindent\textbf{Hypothesis:} \textit{Everything can be found inside a shopping mall.}
\begin{lstlisting}[language=]
(p / possible-01
      :ARG1 (f / find-01
            :ARG1 (e / everything)
            :location (ii / inside
                  :op1 (m / mall
                        :purpose (s / shop-01)))))
\end{lstlisting}
\end{itemize}




\section{5-Shot Prompting Result}
\label{5shot}

We report the results of our 5-shot prompting experiments in \Cref{tab:five-shot-results}.

\begin{table}[h]
\small
\centering
\begin{tabular}{l|cc}
\toprule
Prompt & MNLI & HANS \\
\midrule
Text only &0.89 & 0.82\\
\quad +AMR &0.75 &0.88 \\
AMR only &0.69 & 0.67\\
\bottomrule
\end{tabular}
\caption{Performance comparison with and without AMR on HANS and MNLI test sets in the LLM five-shot prompting setting.}
\label{tab:five-shot-results}
\end{table}

\end{document}